\documentclass[preprint]{elsarticle}
\journal{Pattern Recognition}

\usepackage{lineno}
\modulolinenumbers[5]

\usepackage{amsmath}
\usepackage{amsfonts} 
\usepackage{bbm}

\usepackage{hyperref}
\hypersetup{pdfauthor=author}
\usepackage{graphicx}
\usepackage{multirow}
\usepackage{makecell}
\setcellgapes{5pt}

\usepackage[utf8]{inputenc}
\usepackage[T1]{fontenc}

\usepackage{caption}
\usepackage{subcaption}

\usepackage{subfiles} 
\newcommand{\argmax}{\mathop{\mathrm{argmax}}} 

\def\xset{$\mathcal{X}\,$}
\def\lbls{$\mathcal{Y}\,$}
\def\latent{\mathcal{Z}}
\def\model{$\mathcal{M}\,$}
\def\encoder{\mathcal{E}}
\def\decoder{\mathcal{D}}
\def\dist{$\mathcal{N}\,$}
\def\flow{$\mathcal{F}\,$}

\def\alphamu{\mu_i = T_{\mu_i}(\v{x}_{1:i-1}) \text{,} \quad \alpha_i = T_{\alpha_i}(\v{x}_{1:i-1})}

\def\r#1{\mathbb{R}^{#1}}

\def\v#1{\mathbf{#1}}
\def\basedist#1{p_{#1}(\v{#1})}

\begin{document}
\begin{frontmatter}

\title{Flow Plugin Network for conditional generation}
\author[pwr]{Patryk Wielopolski\corref{mycorrespondingauthor}}
\ead{patryk.wielopolski@pwr.edu.pl}
\author[tplx]{Michał Koperski}
\ead{michal.koperski@tooploox.com}
\author[pwr,tplx]{Maciej Zięba}
\ead{maciej.zieba@pwr.edu.pl}

\cortext[mycorrespondingauthor]{Corresponding author}
\address[pwr]{Faculty of Information and Communication Technology, Wroclaw University of Science and Technology, Wybrzeze Wyspianskiego 27, 50-370 Wroclaw, Poland}
\address[tplx]{Tooploox Ltd., ul. Teczowa 7, 53-601 Wrocław, Poland}

\begin{abstract}
Generative models have gained many researcher's attention in the last years resulting in models such as StyleGAN for human face generation or PointFlow for 3D point cloud generation. However, by default, we cannot control its sampling process, i.e., we cannot generate a sample with a~specific set of the attributes. The current approach is model retraining with additional inputs and different architecture, which requires time and computational resources. We propose a~novel approach that enables to generate objects with a given set of attributes without retraining the base model. For this purpose, we utilize the normalizing flow models - Conditional Masked Autoregressive Flow and Conditional Real NVP, as a Flow Plugin Network (FPN).
\end{abstract}


\end{frontmatter}

\section{Introduction}

In the last years, generative models have achieved superior performance in object generation with a StyleGAN \cite{stylegan} for human face synthesis as the leading example. The area of generative models is not limited only to images but also for example contains 3D point cloud generation where one of the best known models is PointFlow \cite{pointflow}. 

These models perform extremely well in the case of unconditional generation process, however, by default, we cannot control it, i.e., we are not able to generate a sample with a~specific set of the attributes out-of-the-box. To perform such a conditional generation we need to put an additional effort to create a new model with such a functionality. In the case of images, specific solutions were proposed: Conditional Image Generation, with conditional modification of the well-known unconditional generative models - Conditional Variational Autoencoders \cite{cvae} and Conditional Generative Adversarial Networks \cite{cgan}. These approaches provide a good result in conditional image generation. However, the ideal solution would consist of a robust transformation of the unconditional base model to a conditional one.

\begin{figure}[!t]
	\centering
	\includegraphics[width=\textwidth]{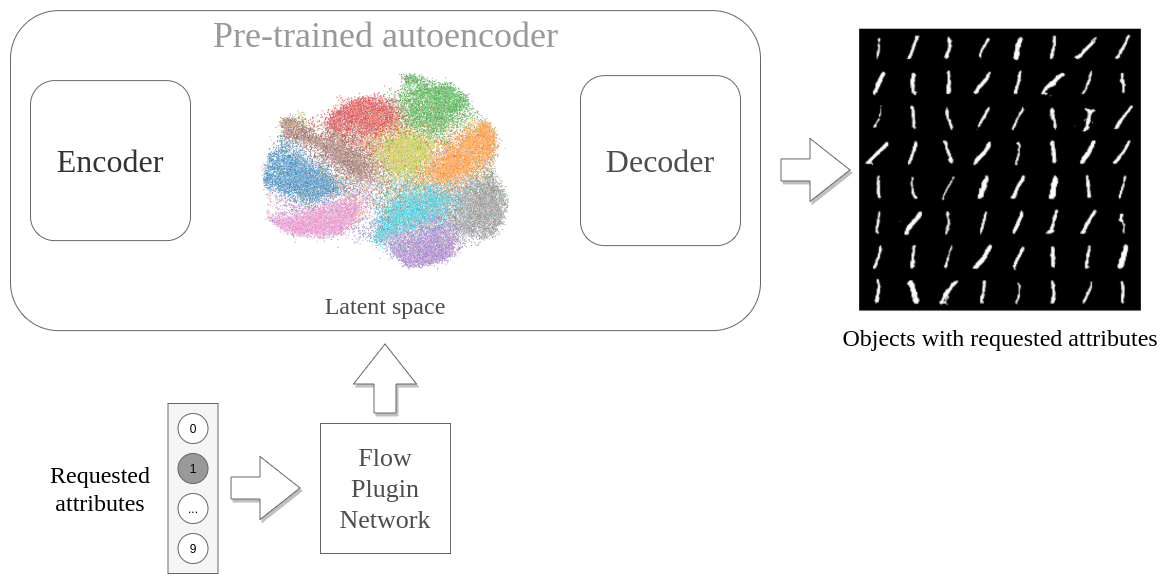}
	\caption{Flow Plugin attached to the latent space of a pre-trained unconditional (both generative and non-generative) autoencoder enables to generate objects with requested attributes.}
	\label{fig:teaser}
\end{figure}

In this work, we propose a novel approach that enables to generate objects with a given set of attributes from already trained unconditional generative model (a base model) without its retraining. We assume that base model has an autoencoder architecture, i.e., it has an encoder and a decoder network which enables to convert image to the latent space representation and invert that transformation respectively (see Figure \ref{fig:teaser}). In our method, we utilize the concept of the plugin network \cite{koperski2020plugin} which task is to incorporate information known during the inference time to the pretrained model and extend that approach to generative models. As the flow plugin model, we will use the conditional normalizing flow models - Conditional Masked Autoregressive Flow and Conditional Real NVP.

We perform several experiments on three datasets: MNIST, ShapeNet, CelebA, which include conditional object generation, attribute manipulation, and classification. The code is publicly available.\footnote{\url{https://github.com/pfilo8/Flow-Plugin-Network-for-conditional-generation}}

Concluding, our contribution are as follows:
\begin{itemize}
	\item we propose a method for conditional object generation from models with an autoencoder architecture, both generative and non-generative;
	\item we show that proposed method could be extended to non-generative variants of the autoencoder models, thus enable to generate objects;
	\item we show that it also could be successfully used for classification tasks or as a tool for attribute manipulation.
\end{itemize}

The rest of the paper is organized as follows, in the next section we describe the theoretical background of the presented methods. Subsequently, in the section \ref{sec:experiments} we present results of the experiments and in the last section we summarize our work and propose directions for future research.

\section{Background}\label{sec:backgorund}

\subsection{Autoencoders}

\begin{figure}[!t]
	\centering
	\includegraphics[width=0.75\textwidth]{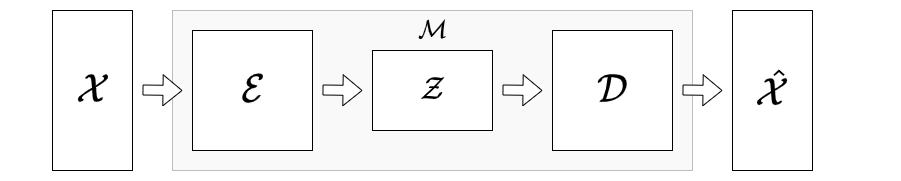}
	\caption{The architecture of autoencoder. It consists of three parts: encoder $\encoder$ - responsible for data encoding, decoder $\decoder$ - responsible for data decoding and latent space $\latent$ - hidden, informative representation of the data. }
	\label{fig:intr-ae}
\end{figure}

Autoencoder \cite{ae} is a neural network which is primarily designed to learn an informative representation of the data by learning to reconstruct its input. Formally, we want to learn the functions $\encoder_{\phi}: \r{n} \rightarrow \r{q}$ and $\decoder_{\theta}: \r{q} \rightarrow \r{n}$ where $\phi$ and $\theta$ are parameters of the encoder and decoder (see Figure \ref{fig:intr-ae}). The model is usually trained by minimizing the reconstruction loss:

\begin{equation}
L_{rec}= \sum_{n=1}^N||\v{x}_n - \decoder_{\theta} \circ \encoder_{\phi}(\v{x}_n)||_2^2, 
\end{equation}
assuming training $\mathcal{X}_N=\{\v{x}_n\}$ is given.

The simple architecture of autoencoder can be enriched by some regularization terms in order to obtain the generative model. For Variational Autoencoder (VAE) \cite{vae} models it is achieved by enforcing the latent representation to be normally distributed using  Kullback-Leibler divergence. For the adversarial autoencoders \cite{aae} the assumed prior on the embeddings is obtained via adversarial training. With such extensions, the model is not only capable to represent the data on low-dimensional manifold but it is also capable to generate samples from an assumed prior in latent space. 

\subsection{Normalizing Flows}

Normalizing flows \cite{papamakarios2021normalizing} represent the group of generative models that can be efficiently trained via direct likelihood estimation. They provide a general way of constructing flexible probability distributions over continuous random variables. Suppose we have a $D$-dimensional real vector $\v{x}$ and we would like to define a joint distribution over $\v{x}$. The main idea of flow-based modeling is to express $\v{x}$, as a transformation $T$ of a real vector $\v{u}$ sampled from a base distribution $\basedist{u}$ with a known density function:

\begin{equation}
\v{x} = T(\v{u}) \quad \text{where} \quad \v{u} \sim \basedist{u} 
\label{eq:flowT}
\end{equation}

The defining property of a flow-based model is that the transformation $T$ must be invertible and both $T$ and $T^{-1}$ must be differentiable which implies that $\v{u}$ must be also a $D$-dimensional vector. Under these conditions, the density of $\v{x}$ is well-defined and can be obtained by a change of variables:
\begin{equation}
\basedist{x} = \basedist{u} |\det{J_{T}(\v{u})}|^{-1} \quad \text{where} \quad \v{u} = T^{-1}(\v{x}) \text{,}
\end{equation}
where the Jacobian $J_{T}(\v{u})$ is the $D \times D $ matrix of all partial derivatives of $T$.

It is common to chain multiple transformations $T_1, \dots, T_K$ to obtain transformation $T = T_K \circ \dots \circ T_1$, where each $T_k$ transforms $\v{z}_{k-1}$ into $\v{z}_{k}$, assuming $\v{z}_0 = \v{u}$ and $\v{z}_K = \v{x}$. The determinant $\det{J_{T}(\v{u})}$ can be expressed as:

\begin{equation}
\det{J_{T}(\v{u})}=\prod_{k=1}^{K} \det{J_{T_k} (\v{z}_{k-1}}) 
\end{equation}



In practice we implement either $T_k$ or $T_k^{-1}$ using a neural network $f_{\phi_k}$ with parameters $\phi_k$. It means that we use $f_{\phi_k}$ to implement $T_k$ to take in $\v{z}_{k-1}$ and output $\v{z}_k$ or conversely, to implement $T_k^{-1}$ which will take $\v{z}_{k}$ and output $\v{z}_{k-1}$. In each case, we need to ensure that the model $f_{\phi_k}$ is invertible and has a tractable Jacobian determinant. Assuming given training data $\mathcal{X}_N=\{\v{x}_n\}$  the parameters $\phi_k$ are estimated in training procedure by minimizing the negative log-likelihood (NLL):

\begin{equation}
-\log{p_{\v{x}}(\v{x}_n)} = 
- \sum_{n=1}^N[\log{p_{\v{u}}(\v{u}_n)} -
\sum_{k=1}^{K} \log{| \det{J_{T_k} (\v{z}_{n,k-1}}) |}]\text{,}
\end{equation}
where $\basedist{u}$ is usually a Gaussian distribution with independent components. The most challenging aspect of training flow-based models is optimisation of $\log{| \det{J_{T_k} (\v{z}_{k-1}}) |}$ that may be challenging to compute for high-dimensional data. That issue is unusually solved by enforcing Jacobian matrix to be triangular, for which the determinant is easy to calculate. In this work, we use two flow-based models, that tackle the problem in different ways: Masked Autoregressive Flow \cite{papamakarios2018masked} and RealNVP \cite{realnvp}.

\subsection{Masked Autoregressive Flow}

Masked Autoregressive Flow \cite{papamakarios2018masked} is an example of the normalizing flow which utilizes the connection with autoregressive models. Practically, it means that it uses the chain rule to decompose any joint density $p(\v{x})$ into a product of one-dimensional conditionals as:

\begin{equation}
p(\v{x}) = \prod_{i=1}^{N}{p(x_i|\v{x}_{1:i-1}}) \text{.}
\end{equation}

Each of the conditional distributions is modelled as single Gaussian, i.e.:
\begin{equation}
p(x_i|\v{x}_{1:i-1}) = \mathcal{N}(x_i|\mu_i, (\exp{\alpha_i)^2}), 
\end{equation}
where $\alphamu$ are unconstrained scalar functions which models mean and log standard deviation of the $i^{\text{th}}$ conditional given all previous variables. They are usually represented by neural networks. 

The model is utilizing the transformation given by \eqref{eq:flowT} independently on a single variable $x_i$ with the following formula:

\begin{equation}
x_i = u_i \exp{(\alpha_i)} + \mu_i \quad \text{where} \quad u_i \sim \mathcal{N}(0,1) \text{.}
\end{equation}

The invert form of the transformation can be easily inverted with the formula:

\begin{equation}
u_i = (x_i - \mu_i)\exp{(-\alpha_i)} \text{.}
\end{equation}

Due to the autoregressive structure of the model, the Jacobian of $T^{-1}$ is triangular by design and its absolute determinant is equal to:
\begin{equation}
|\det{J_{T^{-1}}(\v{x})}| = \exp{\left( -\sum_{i=1}^{N} \alpha_i \right)} \quad \text{where} \quad \alpha_i = T_{\alpha_i}(\v{x}_{1:i-1}) \text{.} 
\end{equation}

\subsection{Real NVP}

Real NVP \cite{realnvp} is an example of the discrete normalizing flow which utilizes the idea of coupling layers. The single coupling layer is defied as:

\begin{align}
\v{x}_{1:d} &= \v{u}_{1:d} \\ 
\v{x}_{d+1:D} &= \v{u}_{d+1:D} \odot \exp{(\v{s}(\v{u}_{1:d}))} + \v{t}(\v{u}_{1:d}) \text{,}
\end{align}
where $\v{s}$ is a scale function, $\v{t}$ is a translation function, where both of them maps data from $\r{d}$ to $\r{D-d}$, and $\odot$ is an element-wise product. This functions are usually represented by deep neural networks and their parameters are optimized during training. The idea of coupling layer is to keep $d$ variables unchanged, and transform the remaining $D-d$ variables by scaling by output of  $\v{s}$ and shifting using the output of $\v{t}$. The motivation behind the transformation that uses the coupling layer is that it is invertible and the Jacobian of this transformation is easy to calculate:

\begin{equation}
\det{J_T(\v{u})} = \exp{\left[ \sum_{j=1}^{D-d}{s(u_{1:d})_j} \right]} \text{.}   
\end{equation}

Real NVP is usually composed of several blocks of coupling layers, where the set of unchanged variables by a previous coupling layer is modified by upcoming one. 

\section{Proposed method} \label{sec:proposed-method}

In this section we present the basic concept of our plugin approach to create a generative model directly from pretrained bottleneck models - Flow Plugin Network (FPN). The idea behind the approach is to incorporate an additional flow model to the latent space of an encoder to enforce generative capabilities without training entire model from scratch. 

The rest of this subsection is organized as follows. First, we are going to provide an overview of the proposed approach. Second, we are going to provide the training details. Third, we will provide a set of practical applications of our model, including conditional generation, classification and attribute manipulation.  

\subsection{Overview}

\begin{figure}[!t]
	\centering
	\includegraphics[width=0.8\linewidth]{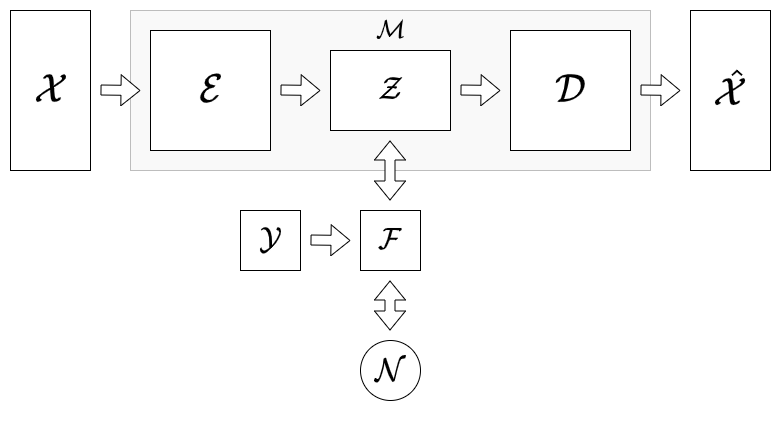}
	\caption{High-level schema of the proposed method.}
	\label{fig:schema-general}
\end{figure}

We consider the base pretrained model \model with an autoencoder architecture (see Figure \ref{fig:intr-ae}), i.e., it has an encoder $\encoder$, a bottleneck with latent space $\latent$, and decoder $\decoder$, which was previously trained on dataset \xset. We also assume that we have a set of attributes \lbls for samples in the dataset \xset. The main idea in the proposed method is to use a conditional normalizing flow model \flow to learn a bidirectional conditional mapping between a simple base distribution \dist (usually standard normal distribution) and a latent space representation which will be conditioned by some attributes (classes) from the set \lbls. This approach will enable us to generate a random sample from the specific region of the latent space which should correspond to object representation with particular set of the attributes. At the last step, using a decoder $\decoder$ we are able to generate an object with a requested set of attributes. The schema of the proposed method can be found in fig. \ref{fig:schema-general}.

\subsection{Training}
\begin{figure}[!t]
	\centering
	\begin{subfigure}[b]{0.45\textwidth}
		\centering
		\includegraphics[width=\textwidth]{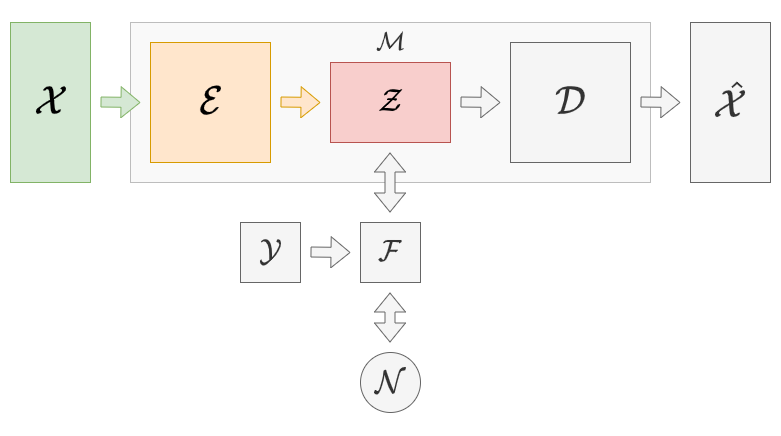}
		\caption{Data encoding.}
	\end{subfigure}
	\hfill
	\begin{subfigure}[b]{0.45\textwidth}
		\centering
		\includegraphics[width=\textwidth]{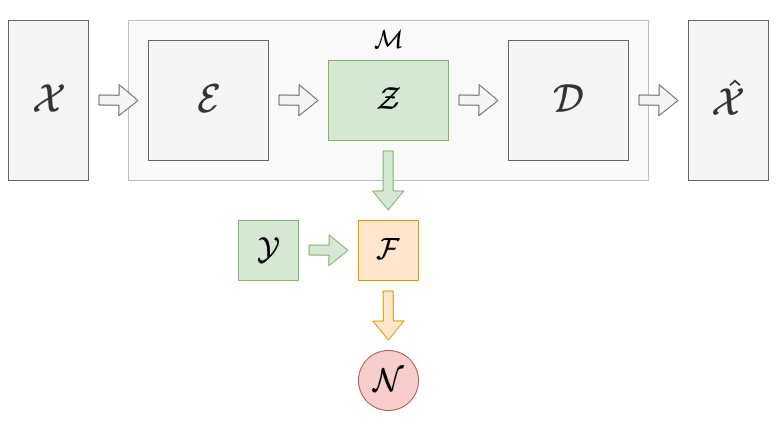}
		\caption{Conditional normalizing flow training.}
	\end{subfigure}
	\caption{Schema of the training procedure. On the left data encoding process where we encode dataset to the latent space representation. On the right procedure of the conditional normalizing flow training where we feed model with vectors from the latent space and the corresponding attributes. Legend: green - input, yellow - intermediate steps, red - output.}
	\label{fig:schema-training}
\end{figure}

We assume that we have a pretrained model \model, samples $\v{x}$ from the dataset \xset and attributes $\v{y}$ from the attribute set \lbls.  Our training goal is to learn a bidirectional mapping between a simple base distribution and a latent space representation, thus we need to have a dataset consisted of pairs $(\v{z}_i, \v{y}_i)$ where $\v{z}_i$ is a latent space representation of the vector $\v{x}_i$. In the first step, we obtain such a dataset by encoding vectors $\v{x}_i$ to vectors $\v{z}_i$ using encoder $\encoder$, as presented in Figure \ref{fig:schema-training} on the left. In the second step, we train conditional normalizing flow \flow using obtained pairs $(\v{z}_i, \v{y}_i)$, by minimizing the conditional negative log-likelihood function:

\begin{equation}
- \sum_{i=1}^{N}{\log{p(\v{z}_i|\v{y}_i)}} = - \sum_{i=1}^{N}{\left( \log{p_u(T^{-1}(\v{z}_i|\v{y}_i))} + \log{|\det{J_{T^{-1}}(\v{z}_i|\v{y}_i)}|} \right)} \text{.}
\end{equation}

It is important to highlight, that weights of the base model \model are freezed during training and only parameters of transformation $T$ are optimized. The form of the transformation $T$ is indicated by the type of the flow and in our experiments we are focused on two types of flows introduced in Section \ref{sec:backgorund}: MAF and RealNVP. 

Because we are focused on constructing model with conditional generative capabilities, we use the conditional version of the flow as a plugin. Practically, it means that the transformation function $T$ takes into account an additional information from conditioning factor (in our case attribute coding $\v{y}$). The form of incorporating conditional information depends strictly on the type of the flow. For MAF the conditioning component is concatenated to the inputs to mean and log standard deviation functions. For RealNVP model the same operation is applied to scaling and shifting functions for each of the coupling layers. The concatenation of the conditioning factor is also performed for invert transformation $T^{-1}$. 

\subsection{Conditional object generation}

\begin{figure}[!t]
	\centering
	\includegraphics[width=0.8\linewidth]{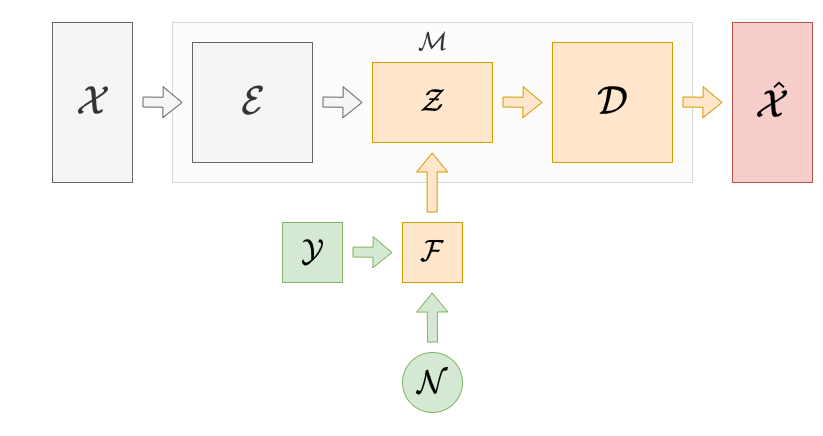}
	\caption{Schema of the conditional object generation. We process sample from the base distribution and requested attributes by the trained conditional normalizing flow and obtain a latent space representation which we decode to the final object using the decoder. Legend: green - input, yellow - intermediate steps, red - output.}
	\label{fig:schema-sampling}
\end{figure}

In this section we are going to present how the proposed approach can be applied for conditional image generation. First, we generate a sample from the base distribution \dist and construct a vector that encodes the attributes $\v{y}$ of the desired object that is going to be generated. Then, the generated sample is passed via the flow conditioned on the vector of attributes embeddings to obtain a vector in the latent space $\latent$ of the base model \model. After that, we process that vector to the decoder and obtain the requested object. The whole process is visualized in Figure \ref{fig:schema-sampling}.

\subsection{Classification} \label{sec:classification}

The proposed approach can be easily applied as generative classification model that utilizes the feature representation $\latent$ of autoencoder. With the conditional normalizing flow \flow we are able to model the conditional probability distribution $P(\v{z}|\v{y})$ where $\v{z}$ is a vector from the latent space and $\v{y}$ is a vector that represents specific class. Assuming that a set $\v{Y}$ is a set of all possible classes we are able to use Bayes rule to calculate predictive distribution $P(\v{y}|\v{z})$, which is probability of specific class $\v{y}$ given the vector of the latent space representation $\v{z}$:

\begin{equation}
P(\v{y}|\v{z}) = \frac{P(\v{z}|\v{y})P(\v{y})}{P(\v{z})} \text{,}
\end{equation}
where $P(\v{y})$ is a assumed class prior and
\begin{equation}
P(\v{z}) = \sum_{\v{y} \in \v{Y}} P(\v{z}|\v{y})P(\v{y}) \text{.}
\end{equation}

As a consequence, we can omit denominator as it is a normalizing factor and perform classification according to the rule given by formula:

\begin{equation}
\hat{\v{y}}  = \argmax_{\v{y} \in \v{Y}} P(\v{z}|\v{y})P(\v{y}) \text{.}
\end{equation}

\subsection{Attribute manipulation} \label{sec:attribute-manipulation}
\begin{figure}[!t]
	\centering
	\begin{subfigure}[b]{0.45\textwidth}
		\centering
		\includegraphics[width=\textwidth]{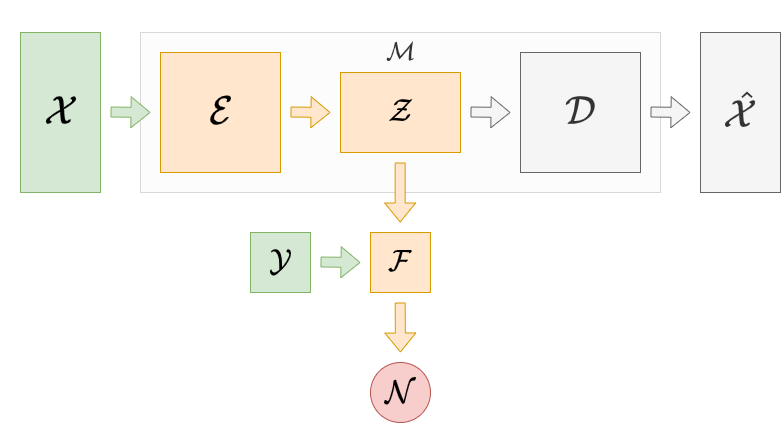}
	\end{subfigure}
	\hfill
	\begin{subfigure}[b]{0.45\textwidth}
		\centering
		\includegraphics[width=\textwidth]{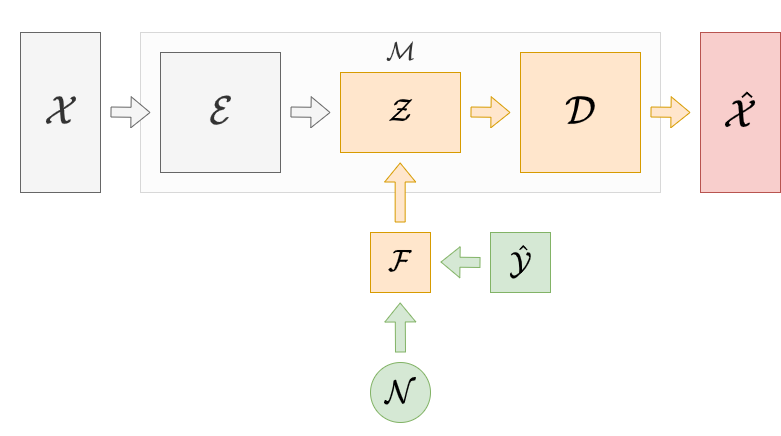}
	\end{subfigure}
	\caption{Schema of the attribute manipulation. On the left we have a process of object encoding to the sample from base distribution. On the right we present process of the sample decoding with changed attributes. Legend: green - input, yellow - intermediate steps, red - output.}
	\label{fig:schema-manipulation}
\end{figure}

We could also use the proposed approach for object's attributes manipulation, e.g., we may want to change the colour of the person's hair on the image. To perform such an operation, firstly, we need to encode an input object to a latent space representation. In the next step, we need to pass obtained sample through the normalizing flow \flow with attributes corresponding to the original image and obtain a sample in the base distribution domain. These two steps are presented in Figure \ref{fig:schema-manipulation} on the left. Subsequently, we need to change the selected attributes. Then, we go back to the latent space representation using the previously obtained sample from the base distribution and the new attributes. Finally, we decode the obtained representation and obtain the final result. These steps are in Figure \ref{fig:schema-manipulation} on the right.

\section{Related works}

The problem of conditional generation is solved using variety of generative approaches. Some of them extend traditional VAE architectures in order to construct the conditional variants of this kind of model for tasks like visual segmentation \cite{cvae}, conditional image \cite{attribute2image} or text generation \cite{cvaenlp}. Conditional variants of GANs (cGANs) \cite{cgan} are also popular for class-specific image \cite{classGAN} or point cloud \cite{points2pix} generation. Conditioning mechanisms are also implemented in some variants of flow models including Masked Autoregressive Flows (MAF) \cite{papamakarios2018masked}, RealNVP \cite{realnvp} and continuous normalizing flows \cite{grathwohl2018ffjord}. The conditional flows are successively applied to semi-supervised classification \cite{atanov2019semi}, super-resolution generation \cite{lugmayr2020srflow}, noise modelling \cite{abdelhamed2019noise}, structured sequence prediction \cite{bhattacharyya2019conditional} or 3D points generation \cite{pumarola2020c}. The presented methods achieve outstanding results in solving variety of tasks. However, they are trained in end-to-end fashion and any changes in the model require very expensive training procedures from scratch. 

That issue is tackled by plugin models introduced in \cite{koperski2020plugin}. The idea behind plugins is to incorporate an additional information, so-called partial evidence, to the trained base model without modifying the parameters. It can be achieved by adding small networks named plugins to the intermediate layers of the network. The goal of these modules is to incorporate additional signal, i.e., information about known labels, into the inference procedure and adjust the predicted output accordingly. Plugin models were successively applied to the classification and segmentation problems. The idea of incorporating an additional conditional flow model to a pretrained generative model was introduced in StyleFlow model \cite{abdal2020styleflow}. This approach was designed for StyleGAN \cite{stylegan}, the state-of-the-art model for generating face images. MSP \cite{li2020latent} utilizes manipulation of VAE latent space via matrix subspace projection in order to enforce desired features on images. 

Comparing to the reference approaches, we present a general framework of utilizing conditional flow for various tasks including conditional image generation, attribute manipulation, and classification. Our model can be applied to any base autoencoder with bottleneck latent space without any additional requirements. 
\section{Experiments} \label{sec:experiments}
In this section, we describe the performed experiments. We evaluate the proposed method on two tasks: conditional object generation and classification. 

In the experiments, we use three different datasets: MNIST~\cite{mnist} -- images of handwritten digits; ShapeNet~\cite{shapenet} -- 3D point clouds of the 55 different objects such as airplanes, cars, laptops; CelebA~\cite{celeba} -- large-scale face attributes dataset with more than 200,000 celebrity images, each with 40 attribute annotations.

For the latent space visualization, we use UMAP \cite{mcinnes2020umap} algorithm as it is one of the fastest dimension reduction algorithm for large datasets. Additionally, UMAP tends to better preserve the global structure of the data, which is important in our case.

For the implementation purpose, we used \textit{nflows} \cite{nflows} in the context of normalizing flows and PyTorch-VAE \cite{Subramanian2020} for Variational Autoencoder.

\subsection{Conditional object generation}
In this section, we show that the  proposed flow plugin network architecture is able to extend an existing pretrained autoencoder to control the class of the generated object.
It is worth noting that our method can do conditional object generation with both generative and non-generative auto-encoders.

\subsubsection{MNIST dataset}
\label{sec:genMNIST}
In the first experiment, we use MNIST dataset, where we extend Variational Autoencoder with proposed flow plugin architecture to conditionally generate images of digits.

\begin{figure}[!t]
	\centering
	\begin{subfigure}[b]{0.37\textwidth}
		\centering
		\includegraphics[width=\textwidth]{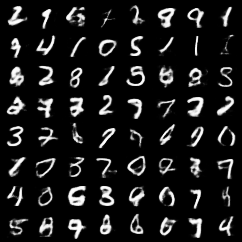}         
	\end{subfigure}
	\hfill
	\begin{subfigure}[b]{0.59\textwidth}
		\centering
		\includegraphics[width=\textwidth]{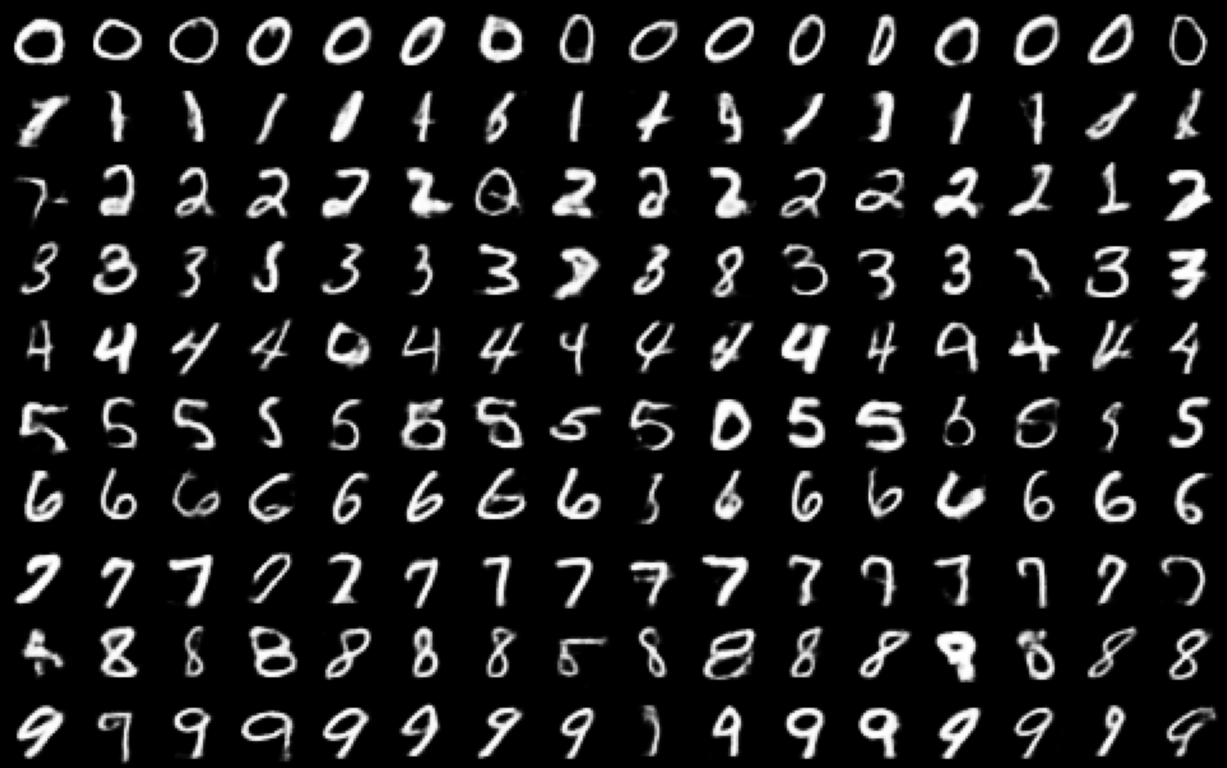}
	\end{subfigure}
	\caption{Conditional object generation on MNIST dataset. On the left side generated samples using ConvVAE model (base model). On the right side generated samples with flow plugin model (Conditional Masked Autoregressive Flow) conditioned on the specific digit class.}
	\label{fig:mnist}
\end{figure}

First, we training Variational Autoencoder as we need some unconditional base generative model from which we want to conditionally generate images of handwritten digits. In our case, it will be a Convolutional Variational Autoencoder (ConvVAE) with 40-dimension latent space. The encoder network was made of 2 blocks consisting of Convolutional Layer, Batch Normalization, LeakyReLU activation function and last Convolutional Layer. The decoder network was made of 3 blocks consisting of Transposed Convolutional Layer, Batch Normalization, ReLU activation function and the last Transposed Convolutional Layer with Sigmoid activation function. The random samples generated from the base model are visualized in the Figure~\ref{fig:mnist} on the left.

To extend the base generative model to conditionally generate images, we train a flow plugin network which will be responsible for bidirectional mapping between the latent space and the base distribution. We use a standard normal distribution and train two flow plugin models: Conditional Masked Autoregressive Flow (C-MAF) and Conditional Real NVP (C-RNVP). The C-MAF was made of 5 layers which each consisted of reverse permutation transformation and a masked affine autoregressive transform implemented using Masked Autoencoder for Distribution Estimation (MADE) with 2 residual blocks. The C-RNVP was made of 5 layers which each used a checkerboard mask and for translation and scale modeling 2 block residual networks were used. In both cases class labels were one hot encoded. 

\begin{figure}[!t]
	\centering
	\includegraphics[width=\linewidth]{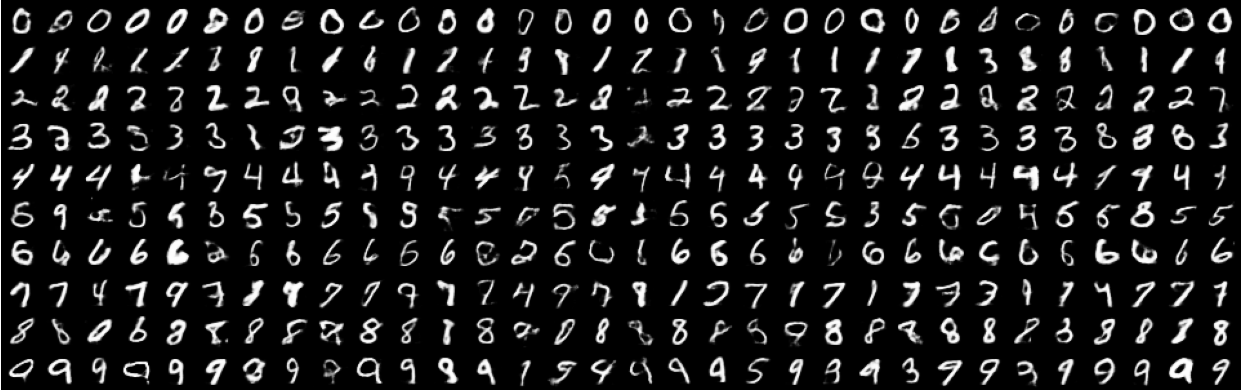}    
	\caption{Conditional object generation on MNIST dataset. Example of generated samples using another flow plugin model (Conditional Real NVP) conditionined on the specific digit class.}
	\label{fig:mnist-sample-rnvp}
\end{figure}

We generate samples from the base model by sampling latent space vectors using the flow plugin model and decoding them by the decoder from the base model. The results are presented in Figure \ref{fig:mnist} for Conditional Masked Autoregressive Flow on the right and in Figure \ref{fig:mnist-sample-rnvp} for Conditional Real NVP. The qualitative results show that our flow plugin network approach can conditionally generate images. Thus we are able to extended the standard ConvVAE model to the new task, and we do not require to change any of its parameters. 
We also observe that the choice of the normalizing flow model used as a flow plugin network did not have much impact on the qualitative results.

\begin{figure}[!t]
	\centering
	\begin{subfigure}[b]{0.48\textwidth}
		\centering
		\includegraphics[width=\textwidth]{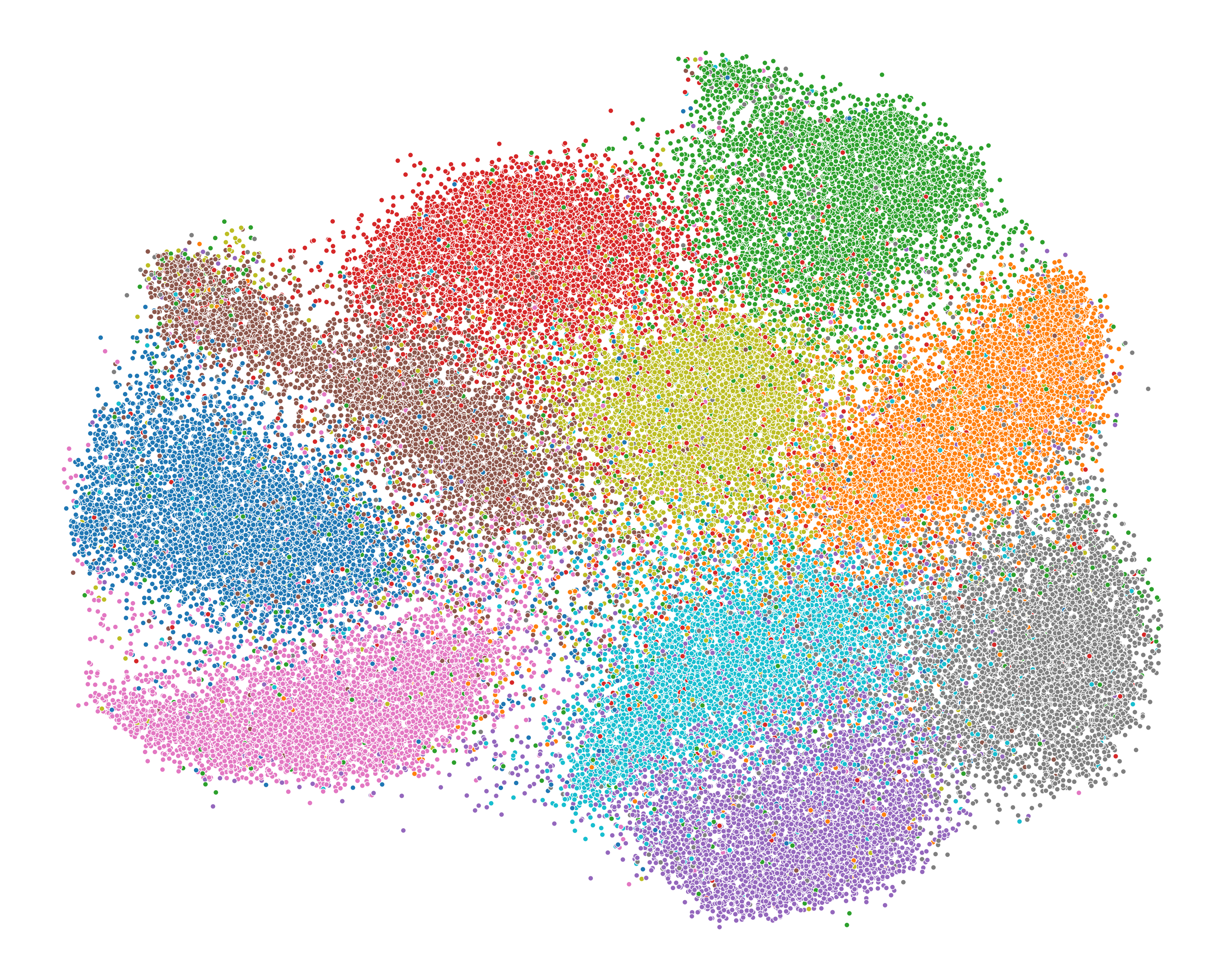}
		\caption{Model}
	\end{subfigure}
	\hfill
	\begin{subfigure}[b]{0.48\textwidth}
		\centering
		\includegraphics[width=\textwidth]{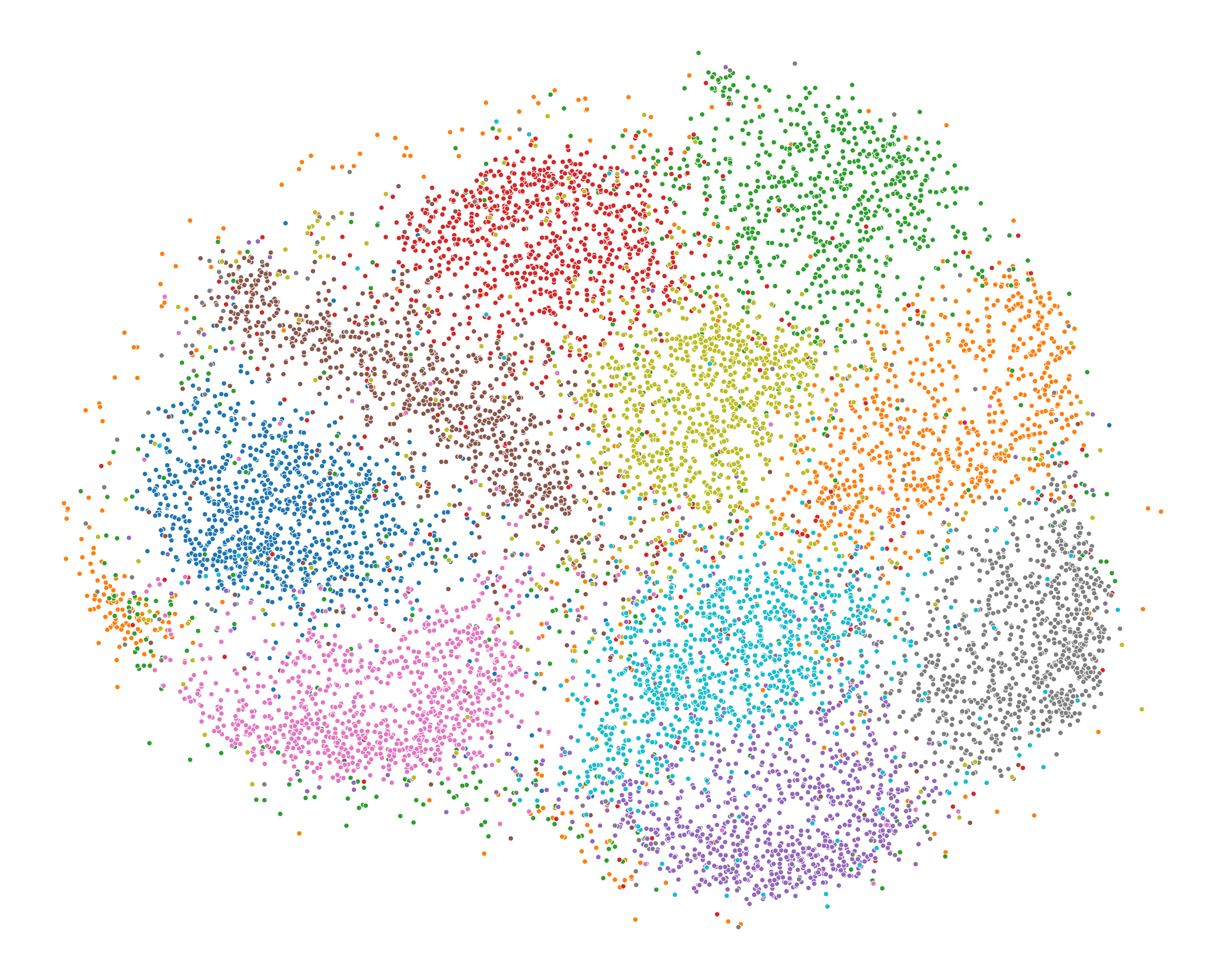}
		\caption{Flow}
	\end{subfigure}
	\caption{Visualization of the latent space of the Autoencoder model and samples generated using Plugin network. Number of samples: 1000}
	\label{fig:mnist-ls-vvae}
\end{figure}

In the second experiment on MNIST dataset we follow similar protocol, but this time we use non-generative autoencoder as a base model. We show that our proposed method is able to extend the base non-generative model to the generative model, and we also allow conditional image generation. As the base model, we train an autoencoder (ConvAE) with the same architecture as in the previously described ConvVAE model.

\begin{figure}[!t]
	\centering
	\begin{subfigure}[b]{0.48\textwidth}
		\centering
		\includegraphics[width=\textwidth]{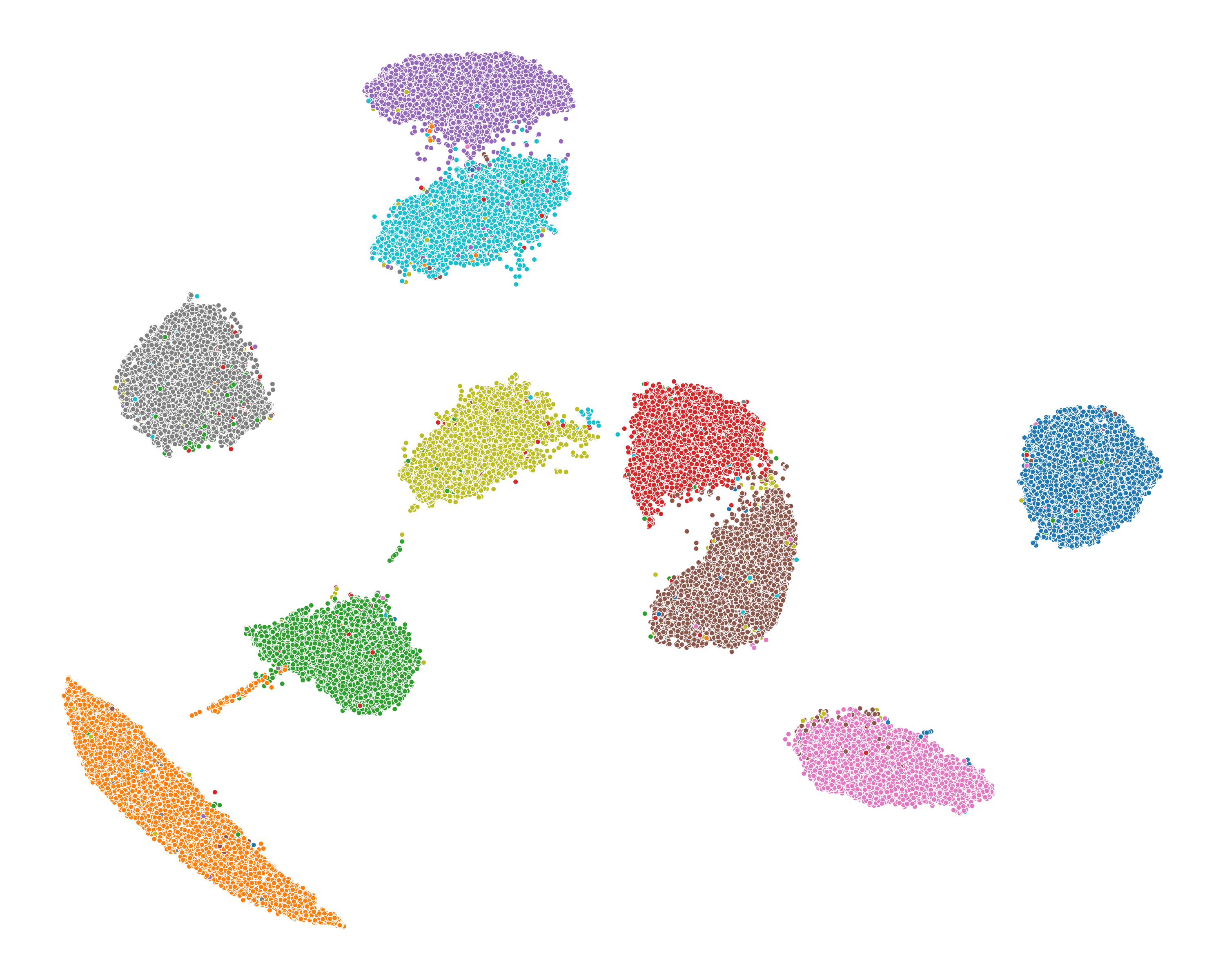}
		\caption{Model}
	\end{subfigure}
	\hfill
	\begin{subfigure}[b]{0.48\textwidth}
		\centering
		\includegraphics[width=\textwidth]{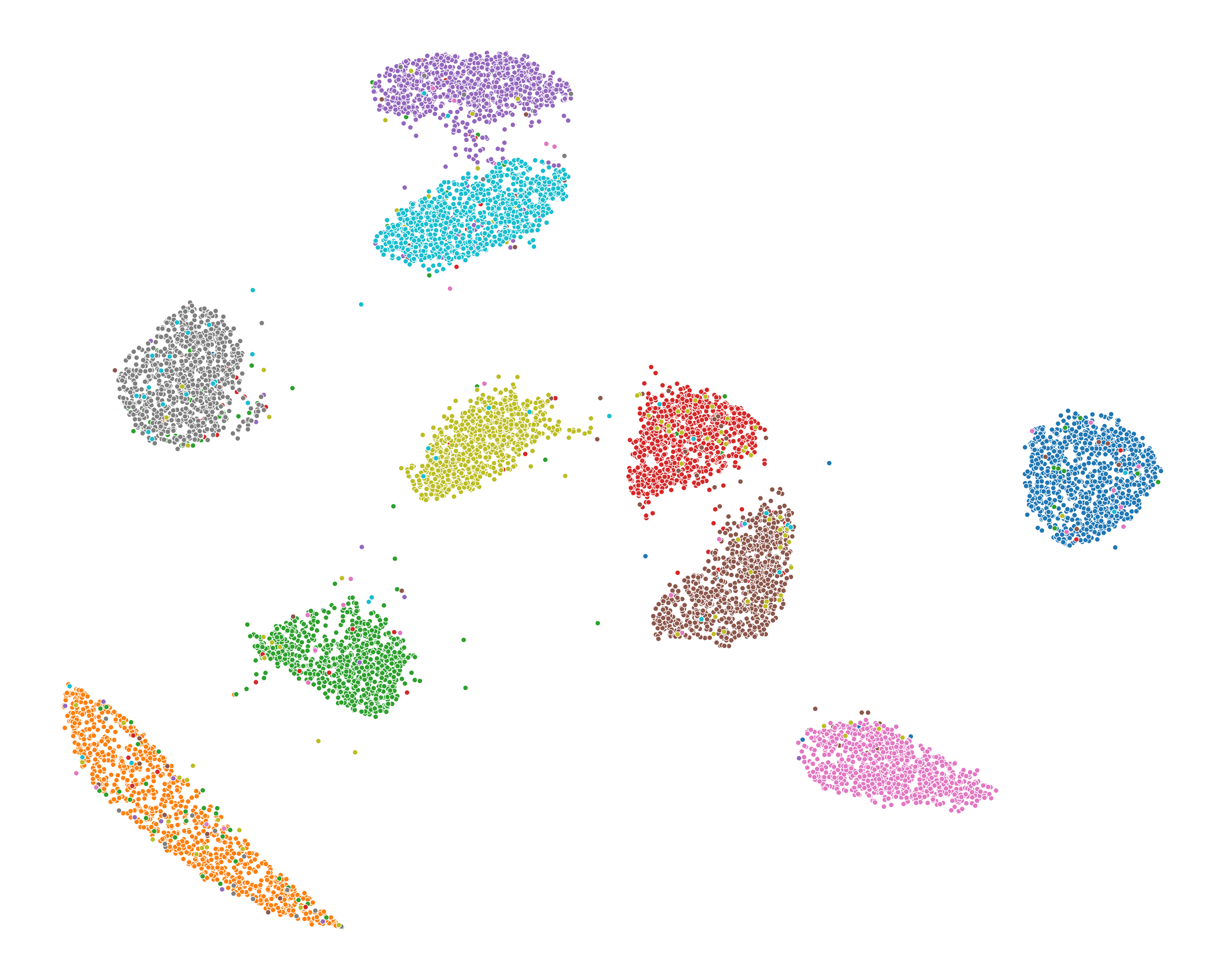}
		\caption{Flow}
	\end{subfigure}
	\caption{Visualization of the latent space of the Autoencoder model and samples generated using Plugin network. Number of samples: 1000}
	\label{fig:mnist-ls-ae}
\end{figure}

Similarly to the previous experiment we trained two flow plugin models:  Conditional Masked Autoregressive Flow (C-MAF), (with 10 layers and 5 residual blocks) and Conditional Real NVP (C-RNVP) (with 10 layers and 5 block residual networks for scale and translation modeling). Moreover, due to the larger values of the obtained latent space, for both models, we have used batch normalization between both layers and blocks.

\begin{figure}[!t]
	\centering
	\begin{subfigure}[b]{0.48\textwidth}
		\centering
		\includegraphics[width=\textwidth]{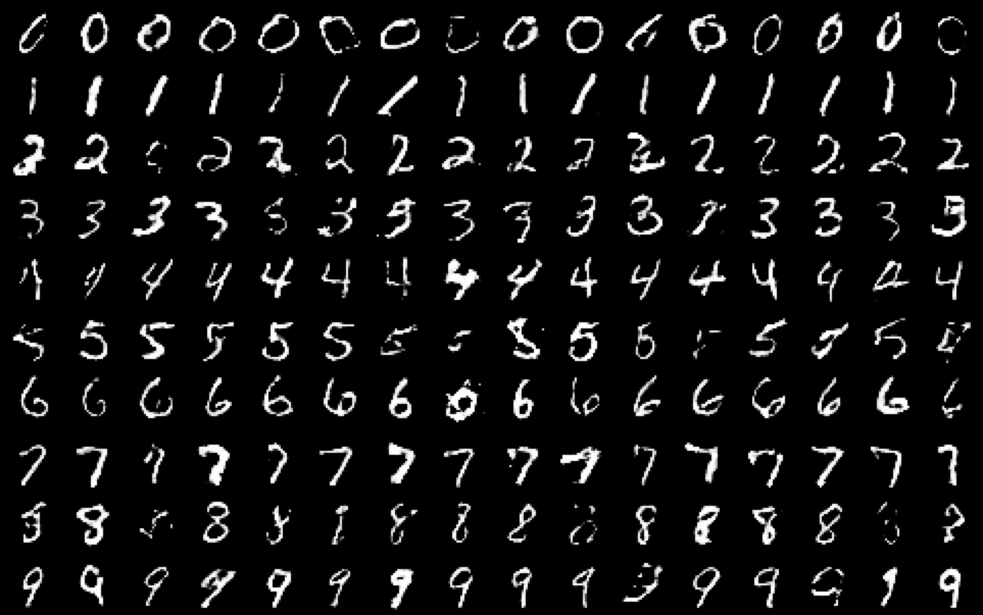}         
	\end{subfigure}
	\hfill
	\begin{subfigure}[b]{0.48\textwidth}
		\centering
		\includegraphics[width=\textwidth]{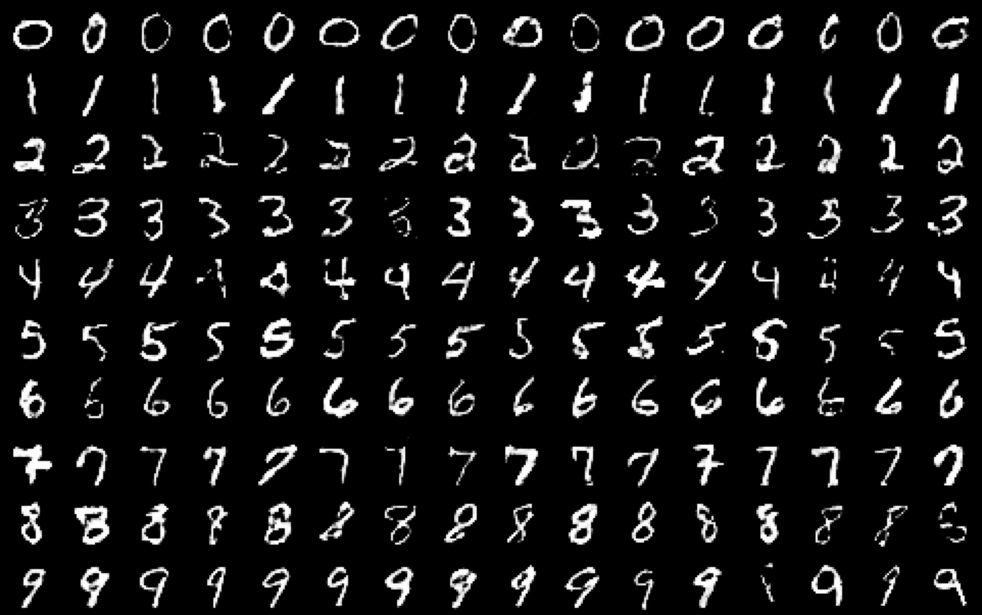}
	\end{subfigure}
	\caption{Results for MNIST experiment with non-generative model. We use the proposed method to extend non-generative model and allow conditional generation of images. Generated samples are from two flow plugin model (C-MAF on left, C-RNVP on right) conditioned on the specific digit class. The base is a ConvAE model.}
	\label{fig:mnist-ext-ae}
\end{figure}

The qualitative results can be found in Figure~\ref{fig:mnist-ext-ae}. First of all, the results are sharper than in the case of ConVAE and this probably is caused by the base model itself and as the autoencoders are less constrained than variational autoencoders and their results can be better. The ConvAE model seems to be a better base model as it achieves better class separation (Figure~\ref{fig:mnist-ls-ae}), comparing to ConvVAE (Figure~\ref{fig:mnist-ls-vvae}).

Our experiments show that the proposed flow plugin architecture can extend both generative and non-generative autoencoders to conditionally generate images of a given class. In addition, our training procedure requires to train only the flow plugin model, while the base model stay unchanged.
This saves a lot of training time as the base model does not need to be retrained. Our experiments also show that the quality of the generated images depends on the quality of the base model. Thus, the possibility to extend both generative and non-generative autoencoders is another important advantage of the proposed approach.

We can also analyze the errors made by our method. Some mistakes can be spotted in Figures~\ref{fig:mnist}, \ref{fig:mnist-sample-rnvp}, e.g. digit 5 is generated as 0 or 4 as 9. It may be caused by the fact that in the latent space some classes are very close to other ones, for example mentioned 5 with 0 or 4 with 9 (see Figure~\ref{fig:mnist-ls-vvae}, left column). Moreover, some areas of the latent space are ambiguous and the flow plugin model may sample observations from this area or the edge of the requested class region. Occurrence of this kind of areas may have two sources: dataset or model. From a dataset perspective, it is not so uncommon to have ambiguous examples and MNIST is not an exception. From a model perspective, our method assumes the pretrained model on which we do not have and do not want to have any influence. However, this observation gives us a guidance that we could obtain better results with models with structured latent space. It is worth noting that the mentioned artifacts do not appear in case of ConvAE, as it provides better class separation (see Figure~\ref{fig:mnist-ls-ae}, left column).

\subsubsection{3D Point Cloud experiments on ShapeNet}
\label{sec:genShapeNet}
The next experiment is a natural extension of the previous one in context of the scale as the dataset has more classes which are also imbalanced and the base model has higher dimensional latent space. Here, we have used PointFlow~\cite{pointflow} as a base model trained on ShapeNet~\cite{shapenet}, the dataset with 55 classes of different objects. The model has an autoencoder-like architecture and operates on a 128 dimensional latent space. In the cited paper, the authors trained and provided seven models: three generative versions for cars, airplanes, and chairs, three non-generative versions for the same three classes, and a non-generative model for the all 55 classes. In our experiments, we use the last one as a base model. We show that we are able to conditionally generate point clouds from the specified class which was not evaluated in the original paper.

\begin{figure}[!t]
	\centering
	\begin{subfigure}[b]{0.48\textwidth}
		\centering
		\includegraphics[width=\textwidth]{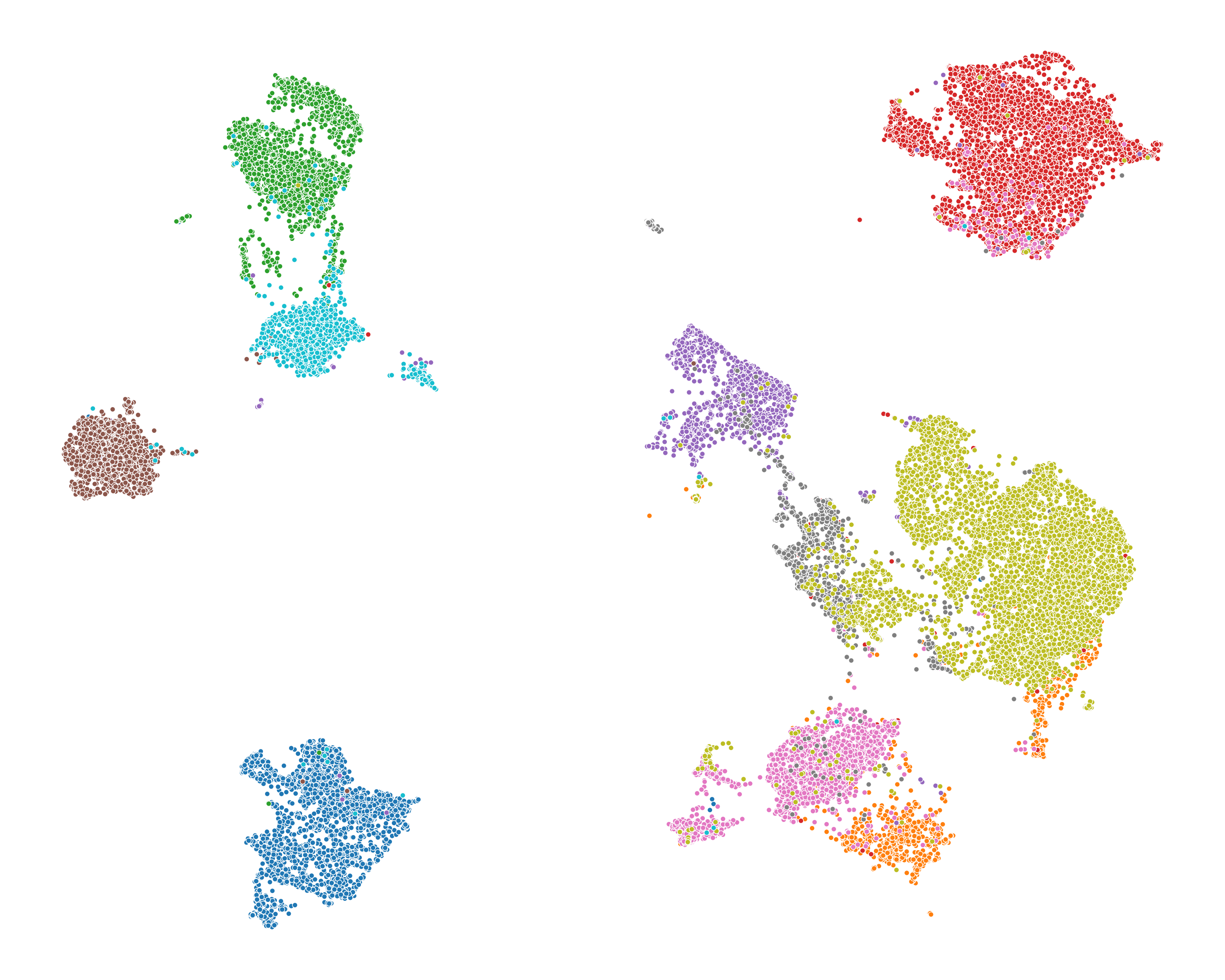}
		\caption{Model}
	\end{subfigure}
	\hfill
	\begin{subfigure}[b]{0.48\textwidth}
		\centering
		\includegraphics[width=\textwidth]{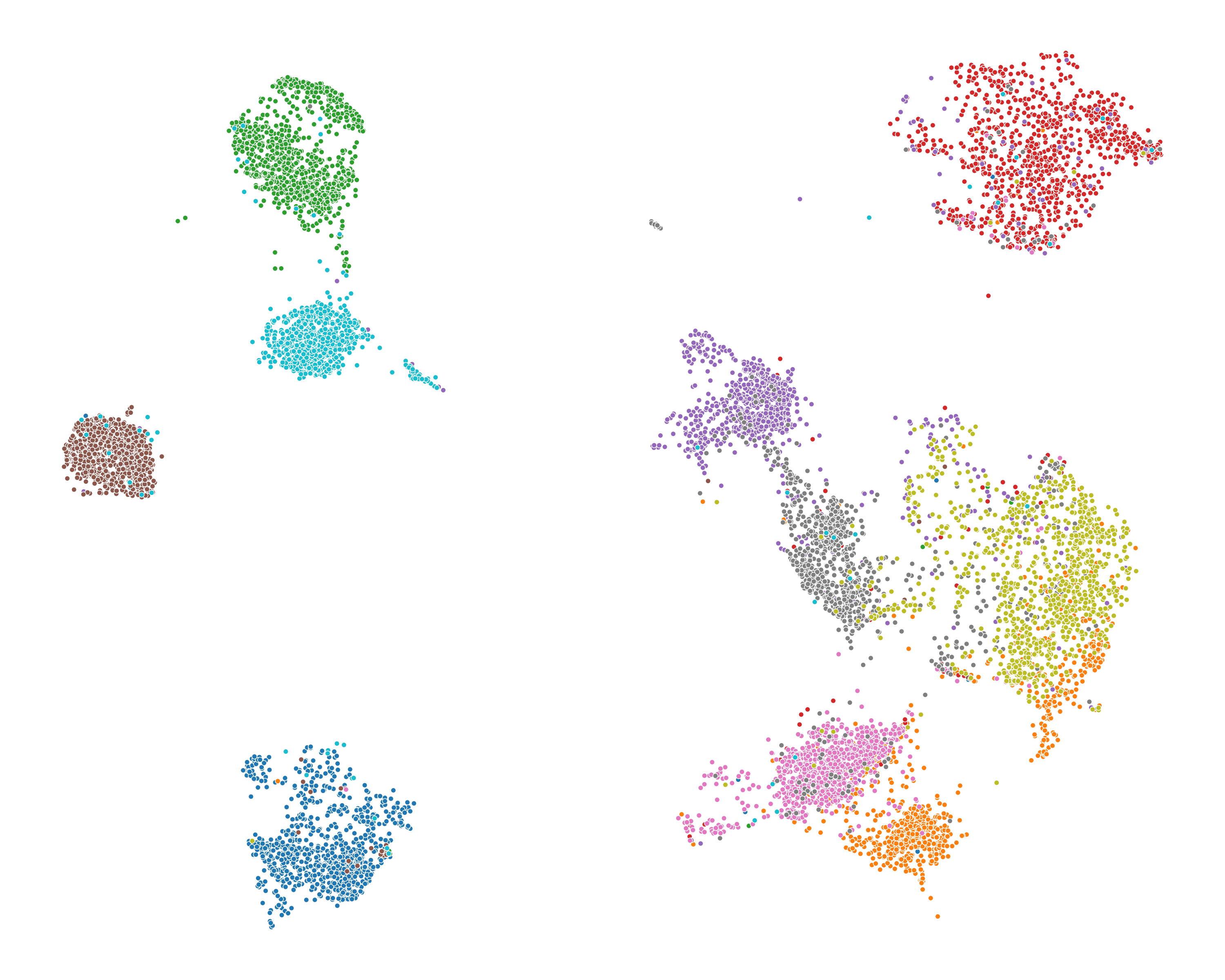}
		\caption{Flow}
	\end{subfigure}
	\caption{Visualization of the latent space of the model and samples generated using Plugin network for 10 most common classes: table, chair, airplane, car, sofa, rifle, lamp, vessel, bench, speaker. Number of samples: 1000}
	\label{fig:ls-10class}
\end{figure}

We trained the Conditional Masked Autoregressive Flow as the flow plugin model, in the exact same manner as in the previous experiment. The latent space of the base model is shown in the left column in~Figure~\ref{fig:ls-10class}. We can observe that the base model captures the class separation, but it does not have a mechanism to conditionally generate the objects. Thus, our approach allows to extend such a model.

\begin{figure}[!t]
	\centering
	\includegraphics[width=\textwidth]{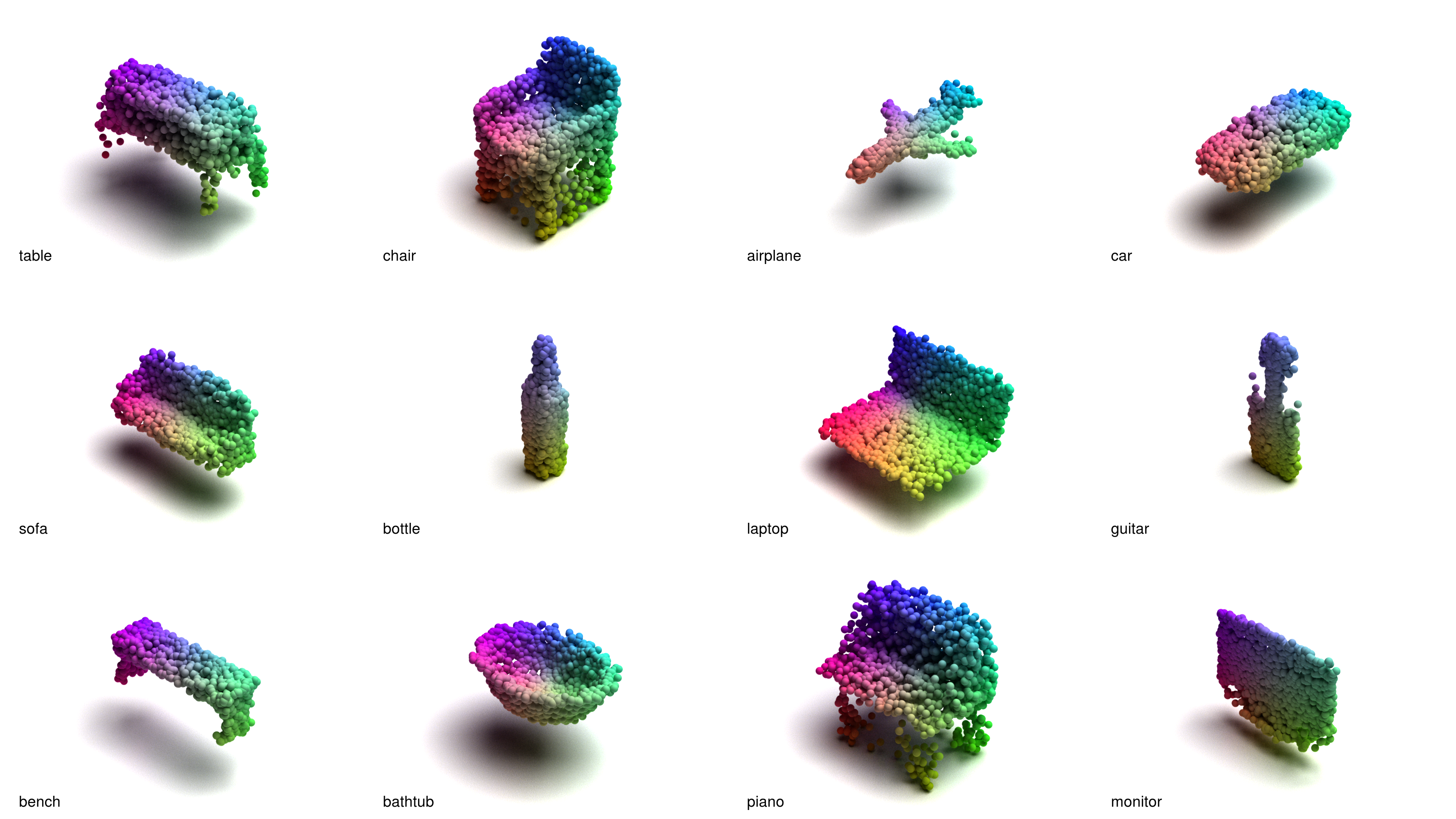}
	\caption{Conditionally generated samples from our flow plugin model. The PointFlow model serves as a base model, while Conditional Masked Autoregressive Model is used as a flow plugin model. Point clouds were conditioned by classes visible under each point cloud.}
	\label{fig:pointflow-flow}
\end{figure}

The conditionally generated results for selected classes from ShapeNet can be found in Figure~\ref{fig:pointflow-flow}. The results show that our method scales to the more complicated case of point cloud generation. We are able to conditionally generate 3D objects from non-generative autoencoder for 55 classes.

Further analysis of the generated object shows a room for improvement in terms of details of the generated objects. The quality is an outcome of couple of factors: 1) quality of the base model, 2) size of the latent space 3) flow plugin model quality. The above factors will be addressed in the future work.

\subsubsection{Attribute manipulation on CelebA dataset}
In this section, we use CelebA~\cite{celeba} dataset. This dataset consists of around 200,000 images with 40 attributes describing among others hair, lips, nose, smile. The challenge of this data comes from the highly correlated attributes. 

We trained Variational Autoencoder with Latent Space Factorization via Matrix Subspace Projection \cite{msp} (MSP) as a base model. Such model architecture disentangles the latent space during the training phase of the model by adding an component responsible for that property to the loss function.

The Conditional Masked Autoregressive Flow is trained as a flow plugin model. It contains 10 layers and 10 residual blocks per layer and batch normalization between layers for more stable training. In addition, we used one linear layer for encoding attributes to the context input for the flow which is meant to be responsible for attributes disentanglement.

\begin{figure}[!t]
	\centering
	\begin{subfigure}[b]{0.45\textwidth}
		\centering
		\includegraphics[width=\textwidth]{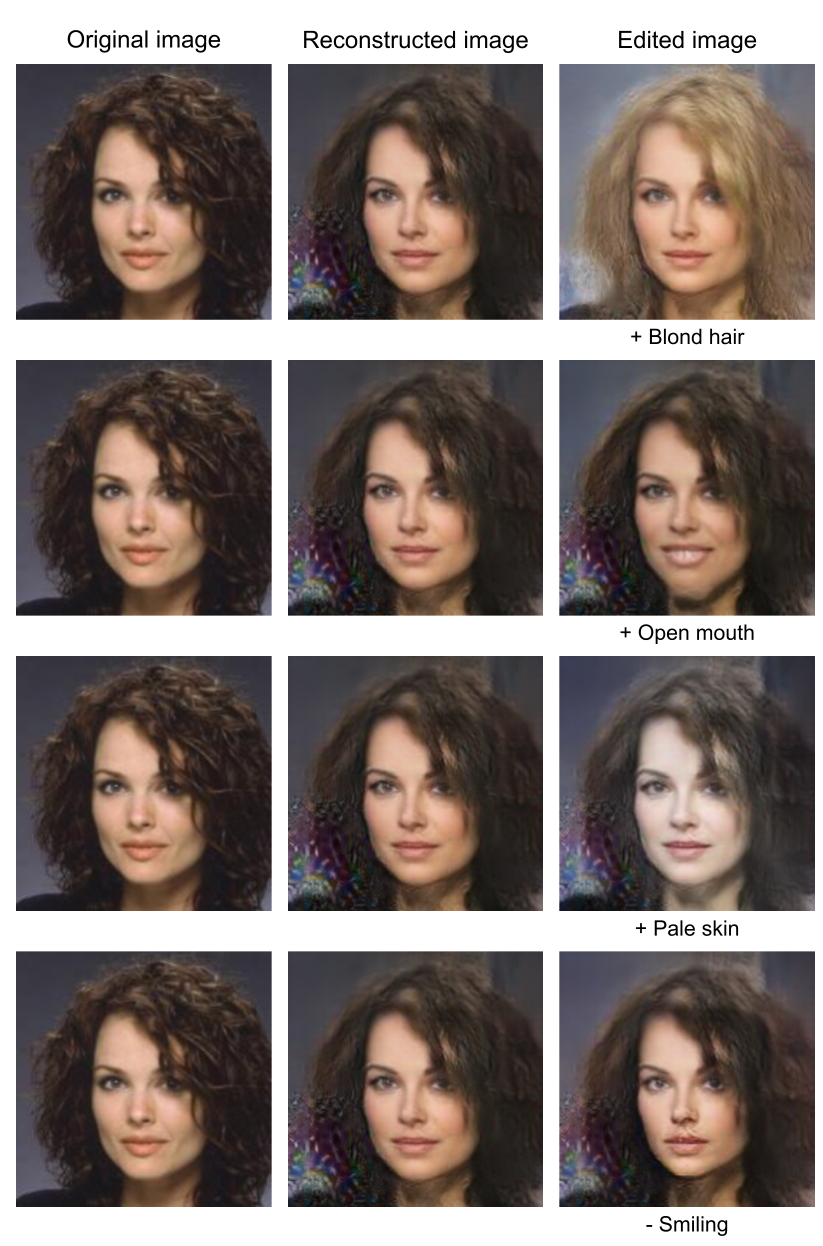}
	\end{subfigure}
	\hfill
	\begin{subfigure}[b]{0.45\textwidth}
		\centering
		\includegraphics[width=\textwidth]{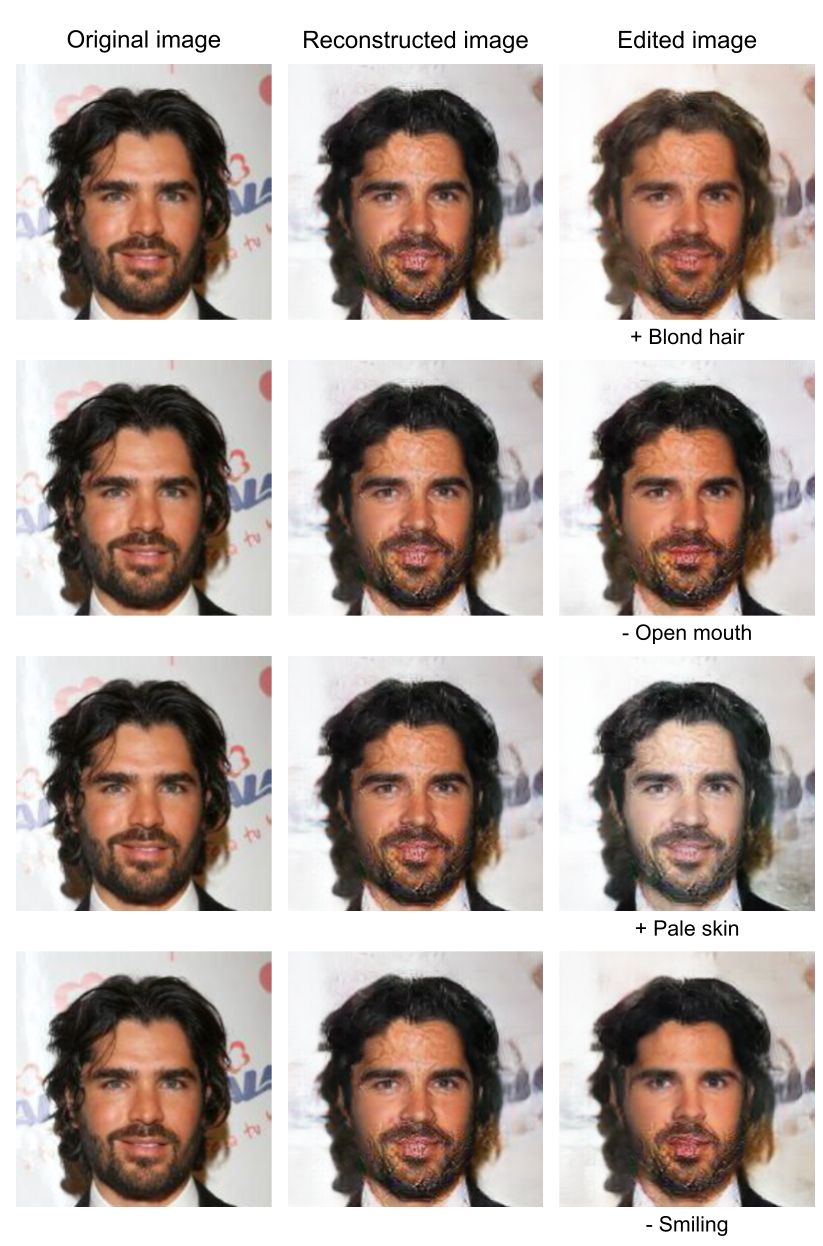}
	\end{subfigure}
	\caption{Results of the feature manipulation experiment on CelebA dataset. Reconstructed image is an output from the MSP base model after encoding and decoding an original image. Edited image is an output of the image manipulation procedure described in Section \ref{sec:attribute-manipulation}. Legend: +/- corresponds to adding or removing particular feature from the conditioning attributes.}
	\label{fig:celeba-fm}
\end{figure}

In this experiment we perform the attribute manipulation experiment as described in \ref{sec:attribute-manipulation}. We pass the original image through base model to obtain the latent representation, then we use our flow plugin model to modify the latent representation to reflect the change of attributes. The results are presented in~Figure~\ref{fig:celeba-fm}. On the left we have the original image, on the middle reconstruction from the base model  and on the right image with changed attributes. We can notice that the base model introduces some artifacts in the reconstruction. The same artifacts are visible in the images generated with our method, as our method is using the base model decoder to generate the image. However, more importantly, the qualitative results show that our proposed method can control the attributes of the reconstructed image. Our model is able to do it without the need of retraining or modifying the base model in any way.

\subsection{Classification}
In this section, we show results of the classification experiment in which we use the proposed flow plugin model as a classifier as described in Section \ref{sec:classification}. It is a very interesting property of our model, as we can extend either generative or non-generative models (trained in an unsupervised way) to classification models.
In our experiments, we compare our method with other classification baselines trained in a supervised way. Specifically, we compare our method to: Logistic Regression, SVM with linear kernel, SVM with RBF kernel, and AutoML. 

\subsubsection{MNIST}
To classify images from MNIST dataset, we use same base and flow plugin models as described in section~\ref{sec:genMNIST}.
For each machine learning model, we performed grid search for the best regularization parameter using 5-fold cross validation. 

The accuracy assessment obtained by all models is presented in Table \ref{table:mnist}. We can observe that the flow plugin models were the best performing models. Obtaining accuracy on the same level as other trained classification models trained in a supervised way. 

\begin{table}[!ht]
	\centering
	\begin{tabular}{|l|l|l|}
		\hline
		\multicolumn{3}{|c|}{MNIST Classification experiment} \\ \hline \hline
		Model                  & Train acc.    & Test acc.    \\ \hline \hline
		Logistic Regression    & 0.8746        & 0.8833       \\ \hline
		Linear SVM             & 0.909         & 0.9164       \\ \hline
		RBF SVM                & 1.0           & 0.9603       \\ \hline
		AutoML                 & 1.0           & 0.9608       \\ \hline
		FPN (Conditional-MAF)                  & 0.9771        & \textbf{0.9677}       \\ \hline
		FPN (Conditional-RealNVP)                 & 0.9740        & 0.9616       \\ \hline
	\end{tabular}
	\caption{MNIST Classification experiment results. Our flow plugin architecture extends unsupervised autoencoders to classification task. The propose method achieves similar level of accuracy as the other classification models trained in a supervised way.}
	\label{table:mnist}
\end{table}

\subsubsection{3D Point Cloud experiments on ShapeNet}
Similarly as in the MNIST experiment, we performed the classification experiment on ShapeNet dataset. We used the same base and flow plugin models as described in section~\ref{sec:genShapeNet}. The obtained results are reported in Table~\ref{table:shapenet}. The dataset is significantly more complicated than MNIST. And although our method does not outperform other models which are trained in a supervised way, we obtain comparable results. It is worth noting that we can perform classification using autoencoder as a base model. More importantly, we do not require to retrain or modify the base model.  

\begin{table}[!ht]
	\centering
	\begin{tabular}{|l|c|c|}
		\hline
		\multicolumn{3}{|c|}{ShapeNet Classification experiment} \\ \hline \hline
		Model                  & Train acc.    & Test acc.    \\ \hline \hline
		Logistic Regression    & 0.9030        & 0.8443       \\ \hline
		Linear SVM             & 0.9104        & 0.8503       \\ \hline
		RBF SVM                & 0.9778        & \textbf{0.8643}       \\ \hline
		AutoML                 & 0.9018        & 0.8435       \\ \hline
		FPN (Conditional-MAF)        & 0.9548        & 0.8404       \\ \hline 
	\end{tabular}
	\caption{ShapeNet Classification experiment results. The proposed method reports comparable results to other supervised classification models.}
	\label{table:shapenet}
\end{table}

\section{Summary}
In this work, we have proposed and successfully tested in experiments a method for conditional object generation based on a model with an autoencoder architecture, both generative such as Variational Autoencoder and non-generative such as standard Autoencoder. In case of non-generative autoencoders, this method make them generative. Moreover, we were able to use a trained flow for classification and attribute manipulation tasks.

In more detail, during experiments, we have shown that using the proposed method, we are able to conditionally generate images and 3D point clouds from generative models such as Variational Autoencoder, Variational Autoencoder with Matrix Subspace Projection and non-generative ones such as Autoencoder or PointFlow. Moreover, we have performed a classification task on MNIST and ShapeNet dataset and compared the results with shallow machine learning models and AutoML approach where on the former dataset we obtained the best results using our model and on the latter worse, however still comparable. Lastly, we have successfully manipulated images of human faces, however, without comparison between existing methods.

Our method could be extended to other generative models, i.e., Generative Adversarial Networks and Normalizing Flows, and only dataset collection is an obstacle for performing such a test which we are planning to do in the future. Especially, the well-known StyleGAN model could be a good reference point for exploration of the Generative Adversarial Networks. Moreover, in this work we have focused only on images and point clouds, however, as this approach should work for any autoencoder model, an interesting future direction may be an application to text or graphs, especially in the context of generation or manipulation. We have provided a comparison between MAF and Real NVP, however, we think that this topic could be much more explored and will be also a part of the future study.

\section{Acknowledgements}

The work of M. Zieba was supported by the National Centre of Science (Poland) Grant No. 2020/37/B/ST6/03463.

\bibliography{main}

\begin{thebibliography}{10}
\expandafter\ifx\csname url\endcsname\relax
  \def\url#1{\texttt{#1}}\fi
\expandafter\ifx\csname urlprefix\endcsname\relax\def\urlprefix{URL }\fi
\expandafter\ifx\csname href\endcsname\relax
  \def\href#1#2{#2} \def\path#1{#1}\fi

\bibitem{stylegan}
T.~Karras, S.~Laine, T.~Aila, A style-based generator architecture for
  generative adversarial networks (2019).
\newblock \href {http://arxiv.org/abs/1812.04948} {\path{arXiv:1812.04948}}.

\bibitem{pointflow}
G.~Yang, X.~Huang, Z.~Hao, M.-Y. Liu, S.~Belongie, B.~Hariharan, Pointflow: 3d
  point cloud generation with continuous normalizing flows, arXiv (2019).

\bibitem{cvae}
K.~Sohn, H.~Lee, X.~Yan,
  \href{https://proceedings.neurips.cc/paper/2015/file/8d55a249e6baa5c06772297520da2051-Paper.pdf}{Learning
  structured output representation using deep conditional generative models}
  (2015).
\newline\urlprefix\url{https://proceedings.neurips.cc/paper/2015/file/8d55a249e6baa5c06772297520da2051-Paper.pdf}

\bibitem{cgan}
M.~Mateos, A.~González, X.~Sevillano, Guiding gans: How to control
  non-conditional pre-trained gans for conditional image generation (2021).
\newblock \href {http://arxiv.org/abs/2101.00990} {\path{arXiv:2101.00990}}.

\bibitem{koperski2020plugin}
M.~Koperski, T.~Konopczynski, R.~Nowak, P.~Semberecki, T.~Trzcinski, Plugin
  networks for inference under partial evidence (2020).
\newblock \href {http://arxiv.org/abs/1901.00326} {\path{arXiv:1901.00326}}.

\bibitem{ae}
D.~Bank, N.~Koenigstein, R.~Giryes, Autoencoders (2021).
\newblock \href {http://arxiv.org/abs/2003.05991} {\path{arXiv:2003.05991}}.

\bibitem{vae}
D.~P. Kingma, M.~Welling, Auto-encoding variational bayes (2014).
\newblock \href {http://arxiv.org/abs/1312.6114} {\path{arXiv:1312.6114}}.

\bibitem{aae}
A.~Makhzani, J.~Shlens, N.~Jaitly, I.~Goodfellow, B.~Frey, Adversarial
  autoencoders, arXiv preprint arXiv:1511.05644 (2015).

\bibitem{papamakarios2021normalizing}
G.~Papamakarios, E.~Nalisnick, D.~J. Rezende, S.~Mohamed, B.~Lakshminarayanan,
  Normalizing flows for probabilistic modeling and inference (2021).
\newblock \href {http://arxiv.org/abs/1912.02762} {\path{arXiv:1912.02762}}.

\bibitem{papamakarios2018masked}
G.~Papamakarios, T.~Pavlakou, I.~Murray, Masked autoregressive flow for density
  estimation (2018).
\newblock \href {http://arxiv.org/abs/1705.07057} {\path{arXiv:1705.07057}}.

\bibitem{realnvp}
L.~Dinh, J.~Sohl-Dickstein, S.~Bengio, Density estimation using real nvp
  (2017).
\newblock \href {http://arxiv.org/abs/1605.08803} {\path{arXiv:1605.08803}}.

\bibitem{attribute2image}
X.~Yan, J.~Yang, K.~Sohn, H.~Lee, Attribute2image: Conditional image generation
  from visual attributes, in: European Conference on Computer Vision, Springer,
  2016, pp. 776--791.

\bibitem{cvaenlp}
T.~Zhao, R.~Zhao, M.~Eskenazi, Learning discourse-level diversity for neural
  dialog models using conditional variational autoencoders, arXiv preprint
  arXiv:1703.10960 (2017).

\bibitem{classGAN}
A.~Odena, C.~Olah, J.~Shlens, Conditional image synthesis with auxiliary
  classifier gans, in: International conference on machine learning, PMLR,
  2017, pp. 2642--2651.

\bibitem{points2pix}
S.~Milz, M.~Simon, K.~Fischer, M.~P{\"o}pperl, H.-M. Gross, Points2pix: 3d
  point-cloud to image translation using conditional gans, in: German
  Conference on Pattern Recognition, Springer, 2019, pp. 387--400.

\bibitem{grathwohl2018ffjord}
W.~Grathwohl, R.~T. Chen, J.~Bettencourt, I.~Sutskever, D.~Duvenaud, Ffjord:
  Free-form continuous dynamics for scalable reversible generative models,
  arXiv preprint arXiv:1810.01367 (2018).

\bibitem{atanov2019semi}
A.~Atanov, A.~Volokhova, A.~Ashukha, I.~Sosnovik, D.~Vetrov, Semi-conditional
  normalizing flows for semi-supervised learning, arXiv preprint
  arXiv:1905.00505 (2019).

\bibitem{lugmayr2020srflow}
A.~Lugmayr, M.~Danelljan, L.~Van~Gool, R.~Timofte, Srflow: Learning the
  super-resolution space with normalizing flow, in: European Conference on
  Computer Vision, Springer, 2020, pp. 715--732.

\bibitem{abdelhamed2019noise}
A.~Abdelhamed, M.~A. Brubaker, M.~S. Brown, Noise flow: Noise modeling with
  conditional normalizing flows, in: Proceedings of the IEEE/CVF International
  Conference on Computer Vision, 2019, pp. 3165--3173.

\bibitem{bhattacharyya2019conditional}
A.~Bhattacharyya, M.~Hanselmann, M.~Fritz, B.~Schiele, C.-N. Straehle,
  Conditional flow variational autoencoders for structured sequence prediction,
  arXiv preprint arXiv:1908.09008 (2019).

\bibitem{pumarola2020c}
A.~Pumarola, S.~Popov, F.~Moreno-Noguer, V.~Ferrari, C-flow: Conditional
  generative flow models for images and 3d point clouds, in: Proceedings of the
  IEEE/CVF Conference on Computer Vision and Pattern Recognition, 2020, pp.
  7949--7958.

\bibitem{abdal2020styleflow}
R.~Abdal, P.~Zhu, N.~Mitra, P.~Wonka, Styleflow: Attribute-conditioned
  exploration of stylegan-generated images using conditional continuous
  normalizing flows (2020).
\newblock \href {http://arxiv.org/abs/2008.02401} {\path{arXiv:2008.02401}}.

\bibitem{li2020latent}
X.~Li, C.~Lin, R.~Li, C.~Wang, F.~Guerin, Latent space factorisation and
  manipulation via matrix subspace projection, in: International Conference on
  Machine Learning, PMLR, 2020, pp. 5916--5926.

\bibitem{mnist}
Y.~LeCun, C.~Cortes, C.~Burges, Mnist handwritten digit database, ATT Labs
  [Online]. Available: http://yann.lecun.com/exdb/mnist 2 (2010).

\bibitem{shapenet}
A.~X. Chang, T.~Funkhouser, L.~Guibas, P.~Hanrahan, Q.~Huang, Z.~Li,
  S.~Savarese, M.~Savva, S.~Song, H.~Su, J.~Xiao, L.~Yi, F.~Yu, {ShapeNet: An
  Information-Rich 3D Model Repository}, Tech. Rep. arXiv:1512.03012 [cs.GR],
  Stanford University --- Princeton University --- Toyota Technological
  Institute at Chicago (2015).

\bibitem{celeba}
Z.~Liu, P.~Luo, X.~Wang, X.~Tang, Deep learning face attributes in the wild
  (December 2015).

\bibitem{mcinnes2020umap}
L.~McInnes, J.~Healy, J.~Melville, Umap: Uniform manifold approximation and
  projection for dimension reduction (2020).
\newblock \href {http://arxiv.org/abs/1802.03426} {\path{arXiv:1802.03426}}.

\bibitem{nflows}
C.~Durkan, A.~Bekasov, I.~Murray, G.~Papamakarios,
  \href{https://doi.org/10.5281/zenodo.4296287}{{nflows}: normalizing flows in
  {PyTorch}} (Nov. 2020).
\newblock \href {https://doi.org/10.5281/zenodo.4296287}
  {\path{doi:10.5281/zenodo.4296287}}.
\newline\urlprefix\url{https://doi.org/10.5281/zenodo.4296287}

\bibitem{Subramanian2020}
A.~Subramanian, Pytorch-vae, \url{https://github.com/AntixK/PyTorch-VAE}
  (2020).

\bibitem{msp}
X.~Li, C.~Lin, R.~Li, C.~Wang, F.~Guerin, Latent space factorisation and
  manipulation via matrix subspace projection (2020).
\newblock \href {http://arxiv.org/abs/1907.12385} {\path{arXiv:1907.12385}}.

\end{thebibliography}

\end{document}